\documentclass[runningheads]{llncs}

\usepackage[T1]{fontenc}
\usepackage{graphicx}
\usepackage{algorithm}
\usepackage{algcompatible}
\usepackage[algo2e]{algorithm2e} 
\usepackage{amsmath}
\usepackage{color}
\usepackage{esvect}

\RenewDocumentCommand{\vec}{me{_}}{%
  \IfNoValueTF{#2}{\vv{#1}}{\vv*{#1}{\mspace{-2mu}#2}}%
}

\usepackage{latexsym}
\usepackage{amssymb}
\usepackage{multirow}
\usepackage{booktabs}
\begin{document}
\title{TG-PhyNN: An Enhanced Physically-Aware Graph Neural Network framework for forecasting Spatio-Temporal Data}
\titlerunning{TG-PhyNN : Physically-Aware GNN framework}

\author{Zakaria Elabid$\textsuperscript{†}$\inst{1},
Lena Sasal$\textsuperscript{†}$\inst{1}, Daniel Busby\inst{2}, \and Abdenour Hadid\inst{1}}

\def\thefootnote{\textsuperscript{†}}\footnotetext{These authors contributed equally to this work}

\authorrunning{Z. Elabid, L. Sasal et al.}
\institute{Sorbonne Center for Artificial Intelligence, Sorbonne University, Abu Dhabi, UAE \email{\{zakaria.alabid,lena.sasal\}@sorbonne.ae} \and
Total Energies, France. }
\maketitle              % typeset the header of the contribution

\begin{abstract}
Accurately forecasting dynamic processes on graphs, such as traffic flow or disease spread, remains a challenge. While Graph Neural Networks (GNNs) excel at modeling and forecasting spatio-temporal data, they often lack the ability to directly incorporate underlying physical laws. This work presents TG-PhyNN, a novel Temporal Graph Physics-Informed Neural Network framework. TG-PhyNN leverages the power of GNNs for graph-based modeling while simultaneously incorporating physical constraints as a guiding principle during training. This is achieved through a two-step prediction strategy that enables the calculation of physical equation derivatives within the GNN architecture. Our findings demonstrate that TG-PhyNN significantly outperforms traditional forecasting models (e.g., GRU, LSTM, GAT) on real-world spatio-temporal datasets like PedalMe (traffic flow), COVID-19 spread, and Chickenpox outbreaks. These datasets are all governed by well-defined physical principles, which TG-PhyNN effectively exploits to offer more reliable and accurate forecasts in various domains where physical processes govern the dynamics of data. This paves the way for improved forecasting in areas like traffic flow prediction, disease outbreak prediction, and potentially other fields where physics plays a crucial role.

\end{abstract}

\section{Introduction}

In diverse fields ranging from economics and meteorology to newer areas like social and biological systems, the ability to forecast and perform predictive analyses is crucial \cite{surveyForecasting2016}. While traditional forecasting methodologies have predominantly utilized statistical models such as moving average, exponential smoothing and autoregressive models tailored to time-series data within Euclidean spaces \cite{hyndman2008ARIMA}, the rise of deep learning has ushered in significant advancements by exploiting complex, non-linear interdependencies that elude classical models \cite{nbeats,transformers,deepar}.

Graph Neural Networks (GNNs) have emerged as a pivotal advancement in this domain, demonstrating significant utility in analyzing data structured as graphs \cite{scarselli2008graph}. With the advent of GNNs, there has been a notable shift from traditional deep learning methods to applications within non-Euclidean domains \cite{bronstein2017geometric}, which include but are not limited to, social networks, biological ecosystems, and infrastructure networks \cite{zhou2020graph}. This evolution enables the processing of complex relational information that surpasses the capabilities of conventional analytical methods.

While purely data-driven GNNs offer flexibility and have shown significant success in various domains, their application in systems governed by well understood physical laws can sometimes lead to suboptimal or uninterpretable results \cite{liu2024review,zheng2014urban,Jiang_2022Trafficsurvey}. Physics informed neural networks \cite{PINN} inherently incorporate domain knowledge through physical laws. By embedding physical laws into the model structure, the outputs provided are inherently interpretable, which is particularly valuable in scientific and engineering applications where understanding the underlying phenomena is as important as predictive accuracy \cite{interpretability}. Additionally, incorporating physical laws reduces the model’s reliance on massive data sets for training, which is particularly beneficial in fields where data collection is expensive or difficult \cite{kdl}. In complex scenarios where physical laws are not straightforward or only partially understood, physically aware models offer a flexible framework that can integrate partial physical knowledge as soft constraints allowing the model to leverage available physical insights without being overly constrained by them \cite{hybrid}. Physics Informed Graph neural networks have shown some promising results as described in \cite{PGINN2021Han} but often rely on simulated data extracted from differential equation instead of real case scenarios \cite{LIU2023211486,XIANG2024111437}.

This paper introduces a novel methodology, titled TG-PhyNN, that integrates the framework of GNNs with the extrapolation of PINNs, substantially enhancing the performance of standard baseline models such as Graph Convolutional Networks (GCNs), Long Short-Term Memory networks (LSTMs), and Gated Recurrent Units (GRUs). By incorporating Partial Differential Equations (PDEs) into their loss functions, these enhanced models not only outperform their conventional counterparts but also offer greater interpretability and adaptability. Our contributions are detailed as follows:
\begin{itemize}
    \item We present a unique framework that integrates spatio-temporal dynamics with physical laws into several foundational models, augmenting their ability to interpret and adapt to complex and evolving graph structures.
    \item We validate that the incorporation of physics-informed enhancements, notably PDEs in the loss functions, significantly elevates the predictive performance and depth of interpretation of these models.
    \item We provide extensive experimental evidence across diverse datasets, demonstrating our models’ superiority over traditional baselines on multiple performance metrics.
\end{itemize}

%All data and code will be made available on GitHub to promote reproducibility and encourage further developments in the field.
\section{Methodology TG-PhyNN}
%This section provides a foundational overview of the mathematical concepts relevant to our study, specifically focusing on the Liénard-type system and the Lighthill-Whitham-Richards (LWR) model of traffic flow. 
\subsection{Differential Equations}
Nonlinear oscillator systems are ubiquitous in physics and serve as models for a wide array of physical phenomena. In chaos theory and the study of extreme events, Liénard-type nonlinear oscillator systems are particularly prevalent. The Liénard-type oscillator%, when subjected to periodic forcing, 
can exhibit extreme events for a permissible set of parameter values. This system can be described by the equation:
\begin{equation}
    \frac{d^2x}{dt^2}  =-\alpha x \frac{dx}{dt}-\gamma x-\beta x^{3}
\end{equation}
Here, 
$\alpha$
 represents the nonlinear damping, 
$\beta$ denotes the strength of nonlinearity, and 
$\gamma$ is associated with the internal frequency of the autonomous system.

In specific scenarios, such as those involving chaotic datasets, the time series of viral diseases like Chickenpox and Covid-19 can be modeled as extreme events using a Liénard-type equation. This approach allows for capturing the complex, nonlinear dynamics inherent in the spread of these diseases. The set value of parameters \cite{Lienard-extreme} is approximately given by: $\alpha = 0.45$, $\gamma = -0.5$, $\beta = 0.5.$

Traffic flow problems can be effectively modeled using the Lighthill-Whitham-Richards (LWR) equation. This equation can be adapted to represent various scenarios, including the flow of bicycles, vehicles, and pedestrians within graph-like structures such as buildings. It can even potentially be applied to model the density of diseases. In the context of traffic, the LWR equation describes how the density of vehicles, \( p(x,t) \), changes over time and space along a roadway:
\begin{equation}
\frac{\partial p}{\partial t} + \frac{\partial (pv)}{\partial x} = 0 \text{ and } v = v_{\text{max}} \left(1 - \frac{p}{p_{\text{max}}}\right) 
\end{equation}
Here, \( p \) represents the vehicle density, and \( v \) is the velocity, \( p_{\text{max}} \) is the maximum density of vehicles that the roadway can accommodate, and \( v_{\text{max}} \) is the maximum allowable speed within the traffic system. In our application, $v_{max}$ is set to 1 and $p_{max}$ is the maximal value of the training dataset.

In a graph context, multiple scenarios must be considered. To compute the derivative at a node, it is necessary to account for the difference between its flux and the flux of the nodes connected to it. The previous equation can be discretized by following the nodes. For every node \( i \):
\begin{equation}
\frac{dp}{dt} + \frac{1}{N_i} \sum_{j \in N(i)} \frac{vp(i) - vp(j)}{d_{ij}} = 0 \quad 
\end{equation}
where \( N(i) \) is the set of nodes connected to \( i \), $N_i$ is the cardinality of \( N(i) \) and \( d_{ij} \) is the distance between the nodes $i$ and $j$, corresponding in this case to the edge attribute. This formulation allows for the analysis of how traffic density evolves in a graph-like structure, providing insights into congestion and flow dynamics.

\subsection{TG-PhyNN: Temporal Graph Physics Informed Neural Network}
% We consider a spatio-temporal differential operator, denoted by $\mathcal{L}$, that acts on a function $f(x, t)$ to yield $\mathcal{L}(f)(x, t)$. An example of such an operator is the second-order non-linear partial differential equation:

% $$\mathcal{L}_{x,t}(f) = \frac{\partial^2 f}{\partial t^2} + a \frac{\partial f}{\partial t} + b f_t + c \frac{\partial f}{\partial x} + d \frac{\partial^2 f}{\partial x^2} +  \sum_{k = 0:n} e_kf_t^k$$

% where $a$, $b$, $c$, $d$, $e_k$ are constants.
    We consider datasets represented as graphs, denoted by $G = (V, E)$ with nodes $v \in V$ and edges $(v, u) \in E$, where nodes represent spatial positions and each node has a sequence of historical values associated as the input node features, $\mathbf{X} = \{\vec{x}_1, \vec{x}_2, \ldots, \vec{x}_N\}$, where each $\vec{x}_t \in \mathbb{R}^F$. The target variable is denoted by $\mathbf{Y} = \{\vec{x}_{N+1}\}$. We train a GNN to learn complex patterns from these historical values and the physical distribution of the target values. However, unlike Physics-Informed Neural Networks (PINNs), where the governing differential equation dictates the training process, we lack a continuous-time mathematical equation for the observed variables, meaning that time and space cannot be used concretely as a continuous input, therefore compeletely invalidating continuous auto-differentiation method. To overcome the absence of a continuous-time equation, we employ discrete derivatives. For an observed time series $f(t)$ indexed over time $t$, the discrete-time derivative of order one can be written as:
$\frac{df}{dt} \approx \frac{f(t + \delta t) - f(t)}{\delta t}$
where $\delta t$ is the lag or time step between observations.

Similarly, the discrete-space derivative for a graph with $N$ nodes can be approximated as:
\begin{equation}
    \frac{\partial f}{\partial x_i} \approx \frac{1}{N_i} \sum_{j \in \mathcal{N}(i)} (f(i) - f(j))
    \label{spatial_derivatives}
\end{equation}

where: $N_i$ is the number of neighboring nodes of node $i$ in the graph.
$\mathcal{N}(i)$ is the set of neighboring nodes of node $i$.

Furthermore, standard PINN models exploit the differentiability of the Multi-Layer Perceptron (MLP) mapping function to backpropagate the loss function (denoted by $\mathcal{L}$). This assumes the existence of an analytical expression for these derivatives. In mathematical terms, consider an MLP mapping function M with parameters denoted by the weight tensor $w$. The time derivative of `f` can be written as: $\dot{f}_t = \frac{df}{dt} = \frac{\partial M(x,t, w)}{\partial t}$ and similarly $\dot{f}_x = \frac{df}{dx} = \frac{\partial M(x,t, w)}{\partial x}$ 

Classic PINNs then leverage backpropagation to compute the gradient of the loss function with respect to the network parameters `$w$` through these spatio-temporal derivatives. This can be expressed as:

$$\frac{\partial\mathcal{L}}{\partial w} = \frac{\partial \mathcal{L}}{\partial \dot{f}_x} \cdot \frac{\partial \dot{f}_x}{\partial w} = \frac{\partial \mathcal{L}}{\partial \dot{f}_x} \cdot \frac{\partial^2 f}{\partial x \partial w} = \frac{\partial \mathcal{L}}{\partial \dot{f}_x} \cdot \frac{\partial^2 M(x,t, w)}{\partial x \partial w}$$

Where: $\frac{\partial \mathcal{L}}{\partial \dot{f}_x}$ can be calculated directly from the loss and $\frac{\partial^2 M(x,t, w)}{\partial x \partial w}$  is calculated using auto-differentiation.

In our approach however, the absence of a continuous governing equation for the time series data presents a challenge in directly enforcing physical constraints through derivatives. We tackle this issue by employing a two-step prediction strategy. We iterate through the training dataset using a pair of consecutive snapshots $s_t$ and $s_{t+1}$ corresponding to the graph states at timesteps t and t+1. For each snapshot, we extract the node features, the edge connections and the edge attributes which we feed to the GNN. The model's prediction from $s_t$ represents the estimated graph state at the next time step (t+1). Subsequently, we retrieve the features extracted from the next snapshot $s_{t+1}$ which we use to predict the next time step $t+2$. The physics informed loss is computed in regards to the differential equation by substracting the two model predictions with the expected behavior based on the underlying physical principles. For example computing the $1^{st}$ order derivatives at timestep $t+1$ and so forth. For the spatial derivatives, due to the nature of the graph, they can be calculated at timestep t+1 using the predicted value of the node attribute from $s_t$ as expressed in equation \ref{spatial_derivatives}. The losses of the network are expressed as follow:

\begin{equation}
\text{Loss}_{\text{data}} = \sum_{t=1}^{n_{snapshots}}  \sum_{x=1}^{n_{nodes}} \left\| Y_{x,t} - \hat{Y}_{x,t} \right\|^2 
\label{observed_loss}
\end{equation}

\begin{equation}
\text{Loss}_{\text{Phy}} = \sum_{t=1}^{n_{snapshots}} \sum_{x=1}^{n_{nodes}} 
 \|\hat{R}(x,t) \|^2
\label{loss_physics}
\end{equation}
Where $\hat{R}$ corresponds to the discretized form of the differential equation in time and space. The total loss of the network writes:
\begin{equation}
\text{L} =  \lambda_1\text{Loss}_{\text{data}} +\lambda_2\text{Loss}_{\text{Phy}}   
\end{equation}

Where $\lambda_1$ and $\lambda_2$ are tuning hyperparameters chosen by cross-validation.

\subsubsection{Complexity of TG-PhyNN}
The proposed TG-PhyNN method introduces a physical model as an additional constraint in the loss function, ensuring the differential equation holds for each time step and node. This integration does not significantly increase the overall computational complexity of the method. The added complexity arises from the computation of the constraint, which is proportional to the product of the number of time steps and the number of nodes, i.e., $O(N.T)$ where T represents the timesteps and N denotes the nodes. In contrast, when using models such as LSTM or GRU, the complexity includes both the temporal dynamics (sequence length and hidden state size) and the spatial structure (number of nodes) and is given by $O(N.T.h^2)$.
This complexity given for LSTM or GRU can be higher when dealing with models such as GCN or GAT as proven by \cite{blakely2021time}.

\subsubsection{Scalability of TG-PhyNN}
Scalability is a critical consideration for any GNN, especially when designed to handle complex, real-world datasets that may involve large graphs or high-frequency data. The TG-PhyNN framework is designed with scalability in mind. Firstly, we integrate of physical constraints within the TG-PhyNN framework through the loss function rather than the network architecture itself to minimize the overhead introduced by physics-informed components. Secondly, we employ GNN architectures that support sparse data representations and reduce the computational complexity associated with large graphs, where the number of edges and nodes can dramatically increase the required computations. Moreover, the framework is compatible with distributed computing environments, which allows it to scale horizontally by distributing the graph and computations across multiple machines or processing units. 

\begin{figure}[!t]
    \centering
    \caption{Overview of TG-PhyNN Architecture. A GNN is trained to forecast the graph at timesteps t+1 and t+2, which are used to compute the derivatives forming the physical loss. These same predictions are compared to the real values, constituting the labeled loss. The total loss is a linear combination of both losses, which is backpropagated and minimized by the optimizer to enhance the forecasting capabilities.}
    \includegraphics[width=0.7\textwidth]{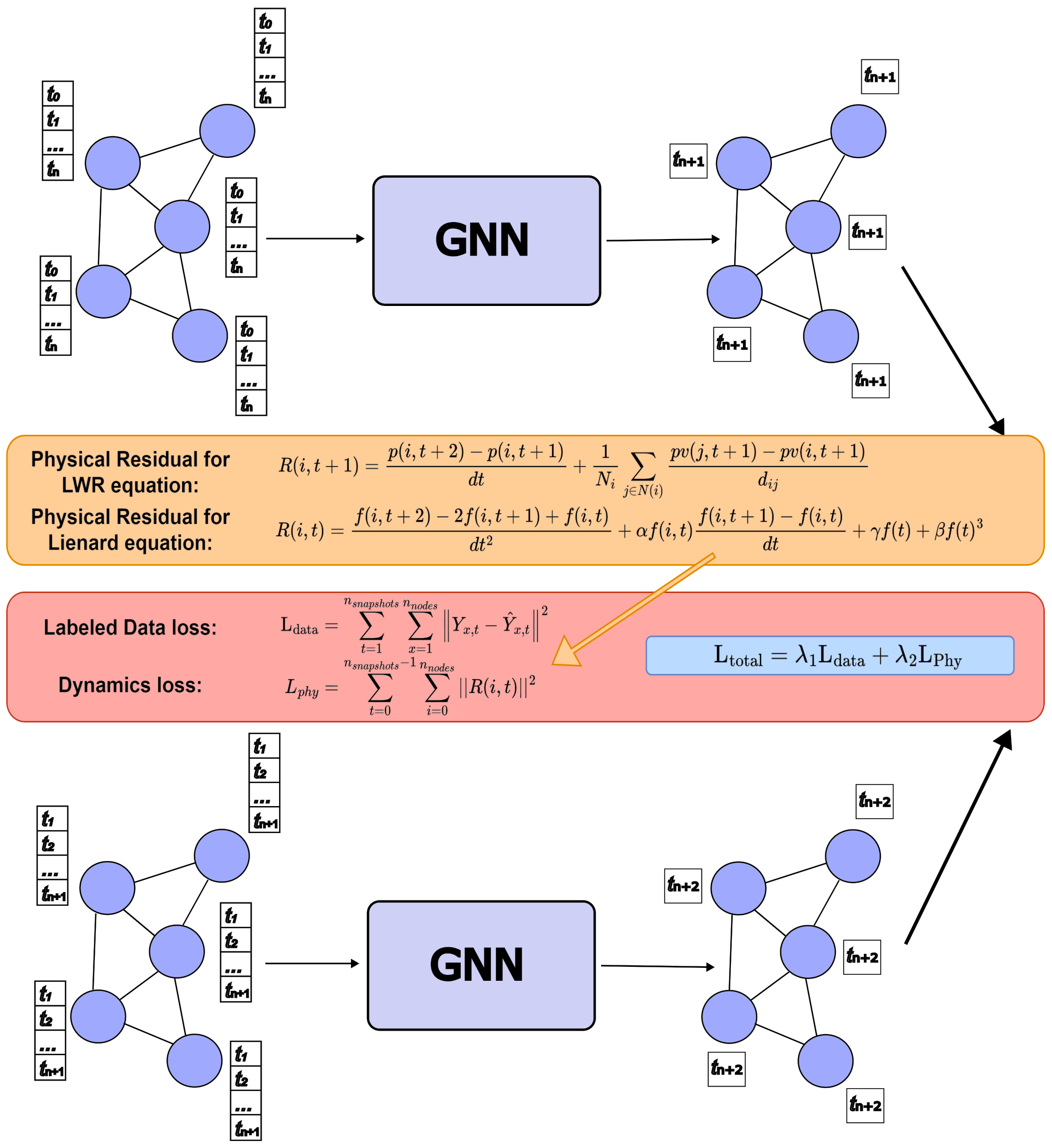}
    
    \label{fig:TG-PhyNN}
\end{figure}

An algorithmic version of the proposed framework is presented in Algorithm \ref{TG-PhyNN} and a schematic diagram showing the various steps of the TG-PhyNN approach is presented in Figure \ref{fig:TG-PhyNN}.

\begin{algorithm}
\caption{\textsc{TG-PhyNN : Physically-Aware GNN framework}}\label{TG-PhyNN}
\KwData{Graph Snapshots $(X, \text{edge\_index}, \text{edge\_attr})$ where $\mathbf{X} = \{\vec{x}_1, \vec{x}_2, \ldots, \vec{x}_t\}$}
\KwResult{Graph Snapshot $(X, \text{edge\_index}, \text{edge\_attr})$ $\mathbf{X} = \{\vec{x}_{t+1}\}$}
\textbf{Networks:} a Graph Neural Network (e.g., GRU, LSTM ...) \\
Initialize networks' parameters: $\theta = (\text{weights and biases})$\;

\For{epoch in range (max\_epochs)} {
    \For{(snapshot, $snapshot_{next}$) in training batch} {
        
        Reset the optimizer gradients\;
        
        Extract features $x$, edge information \text{edge\_index}, \text{edge\_attr} and true labels $y_{t+1}$ from current snapshot \text{snapshot.x}\;
        
        Predict next graph state (t+1) using the model $pred_{t+1}$ = \text{GNN(snapshot.x)}\;
        
        Extract features from next snapshot $snapshot_{next}.x$\;
        
        Predict subsequent graph state (t+2) using the model $pred_{t+2}$ = \text{GNN($snapshot_{next}.x$)}\;
        
        Calculate the observed loss based on predictions $L_{data}(pred_{t+1}, y_{t+1})$\;
        
        Calculate physics-informed loss based on predictions: $L_{phy}(pred_{t+2}, pred_{t+1})$\;
        
        BackPropagate $\lambda_1 L_{data}$ + $\lambda_2 L_{phy}$\ Where $\lambda_1$ and $\lambda_2$ are tuning hyperparameters chosen by cross-validation;
    }
}
\label{algo}
\end{algorithm}

\section{Experimental Setup}

\subsection{Datasets Description}
We employ a diverse array of datasets to evaluate the performance of graph neural network models, offering a robust framework for our experiments. Details of node and edge counts, data collection intervals, and node feature specifics are summarized in Table~\ref{tab:datasets_description}.

\textbf{PedalMe}:
Data from PedalMe Bicycle delivery service in London during 2020-2021 comprises 15 nodes and 225 edges. It includes weekly delivery counts as node features to forecast the next week’s delivery demand.

\textbf{England Covid}:
Daily mobility and COVID-19 case data across England from March to May 2020 form a directed, weighted graph with 129 nodes and 2158 edges. Node features reflect regional COVID-19 statistics to predict the subsequent day's case numbers.

\textbf{ChickenPox}:
Weekly county-level chickenpox case data in Hungary from 2004 to 2014 involves a static graph with 20 nodes and 102 edges. The dataset includes 4 lagged weekly case counts per node, aimed at predicting the next week’s cases.

\begin{table*}[!t]
\centering
\caption{Description of Datasets Used in the Study}
\label{tab:datasets_description}
\begin{tabular}{@{}llllllll@{}}
\toprule
Dataset       & Nodes/Edges & Periodicity & Observations & Past Lag  & Graph Nature \\ \midrule
PedalMe       & 15/225   & Weekly      & 30  &  4 & Static\\
England Covid & 129/2158  & Daily       & 52  &  8 & Dynamic\\
ChickenPox    & 20/102   & Weekly      & 516        & 4 & Static \\ \bottomrule
\end{tabular}
\end{table*}
\subsection{Performance Measures}
To evaluate the accuracy of our models, we utilize the following statistical measures:

 \textbf{Mean Squared Error (MSE)} quantifies the average squared discrepancies between predicted and actual values. It emphasizes the squares of the errors, which disproportionately penalizes larger errors, thereby highlighting significant deviations.
 % The formula for MSE is:
%    \begin{equation}
%        MSE = \frac{1}{n} \sum_{i=1}^{n} (Y_i - \hat{Y}_i)^2
%    \end{equation}

\textbf{Mean Absolute Error (MAE)} computes the average magnitude of the errors in a set of predictions, without considering their direction. It measures the average extent to which the predicted values deviate from the actual values.
%    \begin{equation}
%        MAE = \frac{1}{n} \sum_{i=1}^{n} |Y_i - \hat{Y}_i|
%    \end{equation}

\subsection{Implementation of the TG-PhyNN } %\footnote{All data and code will be made available on GitHub to promote reproducibility and encourage further developments in the field.}
The model is built in Pytorch and utilizes baseline GNN architectures (see 4.2 Baseline Models) from Pytorch Geometric library. Datasets from the PyTorch Geometric Temporal library \cite{rozemberczki2021pytorch} are preprocessed  and split for training (80\%) and testing (20\%). The model employs a two-step prediction strategy within the GNN to incorporate physical constraints as shown in Algorithm \ref{algo}. The model is trained using the Adam optimizer with a learning rate of 0.001 and iterates over 100 epochs. Within each epoch, a batch of graph snapshots is fed into the GNN which performs a two-step prediction to obtain the predicted state at the next two time steps separately. The observed loss (MSE) between predicted and actual values is calculated and the physics-informed loss is computed based on the deviation from the physical equations using the two predicted steps to estimate the temporal derivatives, then the total loss, a weighted sum of the observed and physics-informed losses, is backpropagated through the network to compute gradients. Cross-validation is employed to find the optimal values for weighting the total loss. In our case, we found $\lambda_1$=1 and $\lambda_2$=0.1 to yield the best results.
\section{Results}
\subsection{Data Analysis}
The datasets analyzed PedalMe, ChickenPox, and EnglandCovid underwent several statistical tests %reference statistical tests
to determine their properties \cite{Statisticaltests}. %Stationarity was assessed using the Kwiatkowski–Phillips–Schmidt–Shi (KPSS) test, nonlinearity was examined with Teräsvirta’s neural network test, long-term dependency was evaluated using the Hurst-Exponent, and evidence of chaos was tested with the Lyapunov Exponent test. 
Results indicated that all datasets are non-stationary and non-linear. Additionally, the ChickenPox and EnglandCovid datasets exhibit long-term dependency and chaotic behavior, unlike the PedalMe dataset, which does not display these properties.
\subsection{Baseline Models}

This study compares the proposed TG-PhyNN framework with several established graph-based baseline models, which are crucial for integrating temporal and spatial dynamics in graph-structured data analysis:

\textbf{GRU (Graph GRUs)}: This model extends traditional GRUs to graph data, applying GRU operations on node embeddings updated from neighboring features, thus capturing the temporal dynamics essential for dynamic networks.

\textbf{LSTM (Graph LSTMs)}: Adapting LSTMs for graph data, this model updates node features through embedded LSTM units considering adjacent node features, enabling detection of temporal patterns at multiple scales.

\textbf{GCN (Graph Convolutional Networks)}: GCNs apply a convolution operation tailored for graphs, merging adjacency and node feature matrices to effectively gather and utilize local structural information.

\textbf{GAT (Graph Attention Networks)}: Employing a self-attention mechanism, GATs dynamically weigh the influence of neighboring nodes during feature aggregation, enhancing model performance on complex graph structures.

\textbf{EvolveGCNH (Evolve Graph Convolutional Neural Networks with Highway)}: These networks introduce evolutionary changes to convolutional filters to adapt to dynamic graph structures, supported by highway networks that enhance signal propagation and capture temporal dynamics.

%Each model's ability to handle graph-specific temporal and spatial complexities is critical to our comparative analysis.

\subsection{Results and Discussion}

\begin{table*}[!t]
\centering
\caption{Performance comparison across multiple Real-World datasets: ChickenPox, Covid and PedalMe using baseline models: GRU, LSTM, EvolveGCNH, GCN, GAT and their TG-PhyNN enhanced counterpart. Lower values reflect better performance.}
\label{table:merged_model_performance}
\resizebox{\textwidth}{!}{%
\begin{tabular}{|l|cc|cc|cc|}
\hline
\multirow{2}{*}{\textbf{Model}} &\multicolumn{2}{c|}{\textbf{Chicken Pox}} & \multicolumn{2}{c|}{\textbf{Covid-19}}& \multicolumn{2}{c|}{\textbf{PedalMe}} \\
 & \textbf{MAE} & \textbf{MSE}  & \textbf{MAE} & \textbf{MSE}  & \textbf{MAE} & \textbf{MSE}  \\
\hline
\textbf{GRU}  \cite{seo2016structured}         & 0.6989 & 1.1104  & 0.7031 & 0.7125& 0.8872 & 1.4899    \\
{\textcolor{blue}{\textbf{TG-PhyNN enhanced GRU}}}                   & \textbf{0.6915} & \textbf{1.0969}  & \textbf{0.5850} & \textbf{0.5111}  & \textbf{0.8194}&\textbf{1.3695} \\
\hline
\textbf{LSTM}  \cite{chen2021gclstm}& 0.6511 & 1.0062  & 0.6532 & 0.6044  & 0.7884 & 1.2131   \\
 {\textcolor{blue}{\textbf{TG-PhyNN  enhanced LSTM}}}                   & \textbf{0.6419} & \textbf{0.9967}  & \textbf{0.6500} & \textbf{0.5907}  & \textbf{0.7655}&\textbf{1.0472} \\
\hline
\textbf{EvolveGCNH}  \cite{pareja2020evolvegcn}& 0.6591 & 1.0551  & 0.8941 & 1.0781  & 0.8457 & 1.3510 \\
{\textcolor{blue}{\textbf{TG-PhyNN enhanced EGCNH}}}           &\textbf{0.6484} & \textbf{1.0068}  & \textbf{0.8634} & \textbf{0.9658}  & \textbf{0.7850}&\textbf{1.1147}\\
\hline
\textbf{GCN}   \cite{kipf2017semisupervised}   & 0.6928 & 1.1017 & \textbf{0.6514} & \textbf{0.5791} & \textbf{0.8276} & \textbf{1.2703}   \\
{\textcolor{blue}{\textbf{TG-PhyNN enhanced GCN}}}                    & \textbf{0.6417} & \textbf{0.9931}  & 0.7734 & 0.7677  & 0.9394& 1.5134 \\
\hline
\textbf{GAT} \cite{velickovic2018graph}        & 0.6933 & 1.1008  & 0.7228 & 0.7238  & 1.0948 & 1.8247  \\
{\textcolor{blue}{\textbf{TG-PhyNN enhanced GAT}}}                   & \textbf{0.6482} & \textbf{0.9819}  & \textbf{0.4570} & \textbf{0.3712} & \textbf{1.0128}&\textbf{1.5606}  \\
\hline
\end{tabular}%
}
\end{table*}

The TG-PhyNN models consistently achieved lower MAE and MSE values across all datasets as highlighted in table \ref{table:merged_model_performance}, indicating superior performance. Specifically, the TG-PhyNN enhanced GRU and LSTM models show significant reductions in both MAE and MSE across most datasets, aligning well with the physical dynamics modeled. In the Chicken Pox dataset, the TG-PhyNN version of the GCN model exhibited the highest improvements, with reductions of 7.38\% in MAE and 9.86\% in MSE. The GAT model also showed substantial improvements, with 6.51\% and 10.8\% reductions in MAE and MSE, respectively. For the Covid-19 dataset, the GAT model saw the largest improvements with 36.77\% in MAE and 48.72\% in MSE, while the GRU improved by 16.8\% in MAE and 28.27\% in MSE. However, the TG-PhyNN version of the GCN model underperformed, showing increases in error metrics. In the PedalMe dataset, despite overall positive trends, the TG-PhyNN enhanced models generally performed better, except for the GCN where performance decreased by 13.51\% in MAE and 19.14\% in MSE. These results demonstrate that incorporating physical principles into traditional neural network models through TG-PhyNN enhances their ability to capture and predict complex dynamics, particularly in datasets characterized by non-linear and chaotic behaviors. The TG-PhyNN version of the GCN model also showed less improvement in the Covid-19 dataset. A potential explanation for this anomaly could be the intrinsic limitations of the GCN architecture in adapting to highly dynamic temporal patterns. GCNs are primarily designed to capture spatial dependencies, and the integration of temporal dynamics, particularly in volatile datasets like Covid-19, and thus may require more sophisticated temporal mechanisms than those provided by our current implementation of PDEs in the loss function.

\begin{figure}[!h]
    \centering
    \includegraphics[width=0.7\textwidth]{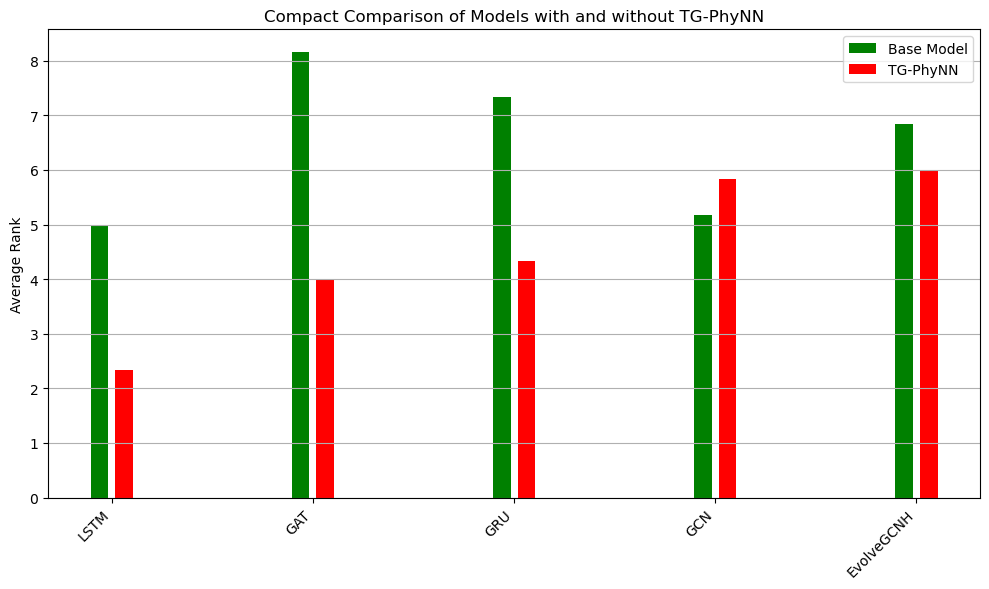}
    \caption{Comparison of average ranking performance across various datasets for base models and models enhanced with TG-PhyNN. The graph illustrates the performance enhancement achieved by incorporating TG-PhyNN into different models such as LSTM, GAT, GRU, GCN, and EvolveGCNH. Each model's performance is ranked based on its predictive accuracy based on both MAE and MSE combining all three datasets, with lower values indicating superior performance.}
    \label{fig:rmse_k_impact}
\end{figure}

Furthermore, no significant improvement was observed with the TG-PhyNN version of the GCN model for the PedalMe dataset. This result suggests that the TG-PhyNN enhancements, particularly the physics-informed components, may not effectively address the complexities inherent in datasets characterized by non-periodic and stochastic variations such as those seen in urban transport dynamics. The irregular patterns in the PedalMe data, influenced by factors like random traffic disruptions, unscheduled urban activities, and varying commuter behaviors, challenge the model’s ability to predict effectively. These findings indicate a need for potentially developing more robust models that can better accommodate the random variability and unpredictability typical of such datasets.
% These results not only validate the efficacy of incorporating physical principles into GNNs but also suggest that TG-PhyNN could be a valuable tool in domains requiring precise forecasting of dynamics governed by known physical or regulatory principles.
The impact of incorporating physical principles into GNNs is further highlighted in figure \ref{fig:rmse_k_impact} where the models are ranked over all datasets based on MAE and MSE. The relevance of ranking in model evaluation lies in its ability to summarize the relative performance of various models across different datasets and metrics, providing an intuitive measure to compare models and simplifying the interpretation of complex performance data. Overall, three of our TG-PhyNN enhanced models, corresponding to LSTM, GAT and GRU achieve the top performance indicating the robustness and practical utility of integrating physical knowledge into data-driven approaches.% but also suggest that TG-PhyNN could be a valuable tool in domains requiring precise forecasting of dynamics governed by known physical or regulatory principles.

\section{Conclusion}
In this paper, we present a novel Temporal Graph Physics-Informed Neural Network framework, titled TG-PhyNN, to incorporate physical constraints into GNN-based forecasting for spatio-temporal data. Our findings showcase significant performance improvements compared to traditional models (GRU, LSTM, GAT) on real-world datasets encompassing traffic flow (PedalMe), disease spread (COVID-19), and outbreaks (Chickenpox). These datasets are governed by well-defined physical principles whose integration leads to more reliable and accurate forecasts across various domains where physical processes dictate data dynamics. All data and code will be made available on GitHub to promote reproducibility and encourage further developments in the field. Looking ahead, the future of TG-PhyNN lies in exploring even more sophisticated methods to integrate physical information. While TG-PhyNN effectively exploits physical principles, some datasets exhibit additional complexities like statistical tendencies and seasonalities that manifest within the governing differential equations. Future work can involve combining the strengths of TG-PhyNN with statistical methods to decompose the network's output.  Promising avenues for achieving this decomposition include Fourier Neural Operators \cite{FNO} or MODWT-enhanced Transformers \cite{sasal2022Wtransformer}, both known for their ability to disentangle different signals within complex data. By incorporating these advanced techniques, we can propel TG-PhyNN towards greater accuracy and broader applicability in real-world forecasting tasks.
% \section*{Aknowledgments}
% The support of TotalEnergies is fully acknowledged. Lena Sasal (PhD Student), Zakaria Elabid (PhD Student) and Abdenour Hadid (Professor, Industry Chair at SCAI Center of Abu Dhabi) are funded by TotalEnergies collaboration agreement with Sorbonne University Abu Dhabi.

\bibliographystyle{splncs04}
\bibliography{Biblio}
\end{document}